\documentclass[10pt,twocolumn,letterpaper]{article}

\usepackage{iccv}
\usepackage{times}
\usepackage{epsfig}
\usepackage{graphicx}
\usepackage{amsmath}
\usepackage{amssymb}
\usepackage{bm,amsfonts}
\usepackage{subcaption}
\newcommand*{\affmark}[1][*]{\textsuperscript{#1}}

\iccvfinalcopy 

\def\iccvPaperID{8019} 

\ificcvfinal\fi

\begin{document}

\title{Weakly-Supervised Photo-realistic Texture Generation for 3D Face Reconstruction}

\author{Xiangnan YIN\affmark[1] \qquad Di HUANG\affmark[2] \qquad Zehua FU\affmark[1] \qquad Yunhong WANG\affmark[2]\qquad Liming CHEN\affmark[1]\\
\affmark[1]Ecole Centrale de Lyon, France \qquad \affmark[2]Beihang University, China\\
{\tt\small \{yin.xiangnan, liming.chen, zehua.fu\}@ec-lyon.fr, \{dhuang, yhwang\}@buaa.edu.cn}
}

\maketitle
\ificcvfinal\thispagestyle{empty}\fi

\begin{abstract}
Although much progress has been made recently in 3D face reconstruction, most previous work has been devoted to predicting accurate and fine-grained 3D shapes. In contrast, relatively little work has focused on generating high-fidelity face textures. Compared with the prosperity of photo-realistic 2D face image generation, high-fidelity 3D face texture generation has yet to be studied. In this paper, we proposed a novel UV map generation model that predicts the UV map from a single face image. The model consists of a UV sampler and a UV generator. By selectively sampling the input face image's pixels and adjusting their relative locations, the UV sampler generates an incomplete UV map that could faithfully reconstruct the original face. Missing textures in the incomplete UV map are further full-filled by the UV generator. The training is based on pseudo ground truth blended by the 3DMM texture and the input face texture, thus weakly supervised. To deal with the artifacts in the imperfect pseudo UV map, multiple partial UV map discriminators are leveraged.
\end{abstract}

\section{Introduction}
3D face reconstruction is an important yet challenging domain in computer vision, aiming to faithfully restore the shape and texture of a face from one or more face images. It has a wide range of applications, such as face recognition, face editing, face animation, and other artistic and entertainment fields. Recently, there has been a surge of interest in single-image based 3D face reconstruction~\cite{deng2019accurate, guo2020towards, zhu2017face,richardson2017learning, guo2018cnn, deng2019accurate, feng2018joint, bai2020deep, tran2018nonlinear}. While most previous work has been devoted to predicting more accurate and detailed 3D shapes, not much work has focused on generating photo-realistic face textures. However, studies ~\cite{masi2019face, hassner2015effective} have shown that the texture plays a far more significant role than that of the shape in face recognition tasks. Thus we can never ignore the importance of the texture in 3D face reconstruction.  
\begin{figure}[t]
\begin{center}
   \includegraphics[width=\linewidth]{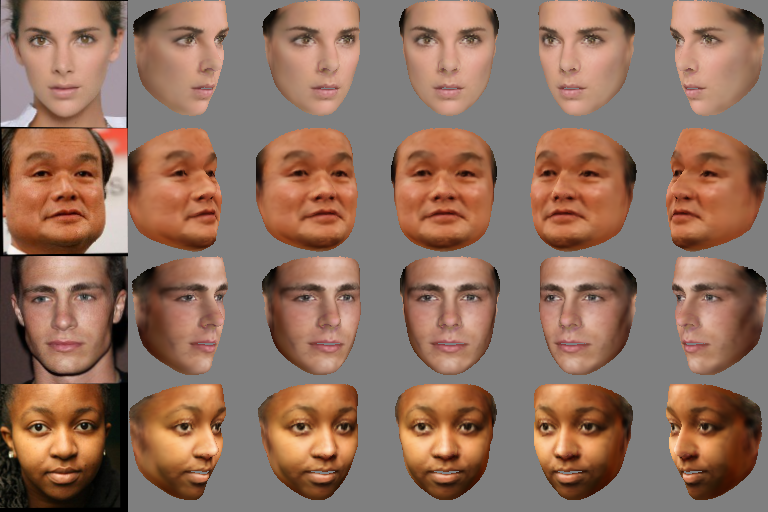}
\end{center}
   \caption{Results of the proposed method. The left column shows the input images. Images on the right are synthesized using the predicted UV-map.}
\label{fig:fig1}
\end{figure}

Existing 3D face texture generation methods can be broadly classified into three categories: texture model-based, image generation-based, and GAN optimization-based. 

\textbf{Texture model-based} Since the 3D Morphable Model (3DMM)~\cite{blanz1999morphable} was proposed, it has been widely used in 3D face reconstruction. The model is a vector basis of the shape and texture learned from a set of 3D face scans. Earlier approaches regress the 3DMM parameters by solving a non-linear optimization problem~\cite{richardson20163d, richardson2017learning, booth20173d}, which is often slow and costly. With the development of Convolutional Neural Networks, recent studies tend to predict the parameters using learning-based methods~\cite{richardson2017learning, guo2018cnn, deng2019accurate}. However, the 3DMM is constructed by a small number of face scans under well-controlled conditions, limiting its diversity to identity, race, age, gender, etc. Besides, due to the linear and low-dimensional nature of the model, it can hardly capture high-frequency details, resulting in blurred textures that are far from satisfactory.

\textbf{Image generation-based} Generative Adversarial Network (GAN) ~\cite{goodfellow2014generative} provides a powerful tool for generating photorealistic images. Since its appearance, numerous image generation methods with stunning results have been proposed. Thanks to various large databases and the highly structured geometry of the human face, 2D face image generation is one of the most prosperous areas ~\cite{cai2019fcsr,huang2018multimodal,pumarola2018ganimation,portenier2018faceshop,choi2018stargan, karras2019style, karras2017progressive}. Influenced by this trend, some recent 3D face reconstruction methods have also leveraged adversarial training to improve the texture quality ~\cite{tran2018nonlinear, deng2018uv, lee2020uncertainty}. However, such an image generation approach is highly dependent on a large 3D face database.~\cite{tran2018nonlinear} is trained on a synthesized 3D face database ~\cite{zhu2016face}, where originally self-occluded textures are obtained by simple interpolation of visible parts, resulting in imperfect generation. ~\cite{deng2018uv, lee2020uncertainty} are trained on a large UV map dataset, which is not publicly available.

\textbf{GAN optimization-based} The traditional yet most powerful GANs are trained to synthesize images from noise vectors~\cite{karras2017progressive,karras2019style,brock2018large}. To leverage the power of a pre-trained GAN, a series of works are established on inverting the image back to a GAN's latent space using optimization-based approaches~\cite{shen2020interpreting,NEURIPS2018_e0ae4561,ma2018invertibility,zhu2016generative}. Similar methods are used to generate the UV map of a face image~\cite{gecer2019ganfit,lee2020styleuv}. First, they train a generator that converts noise vectors into UV maps. Then they directly optimize the latent code to minimize the reconstruction error between the input face image and the image rendered by the generated UV map. Instead of training a UV map generator, ~\cite{gecer2020ostec} first rotates the input image in 3D and optimizes the latent code of the pre-trained StyleGAN to fill in the missing textures, then stitches textures of different view angles by alpha blending to form the final UV map. By far, the optimization-based methods can yield the most realistic face UV maps. Nevertheless, they are usually complex and time-consuming, \textit{e.g.}, GANFIT~\cite{gecer2019ganfit} takes 30 seconds to generate the UV map of an input face, while OSTeC~\cite{gecer2020ostec} takes up to 5 minutes. 

Besides generating a global face texture, we note that a series of pure 2D image generation methods can also synthesize face images of different view angles~\cite{tran2017disentangled,zhou2020rotate,hu2018pose}. However, the generation consistency is poor due to the absence of global consistency constraints and a priori knowledge of the 3D shape. 

In summary, among the current texture generation methods for 3D face reconstruction, those based on texture models cannot yield high-fidelity results due to the model's simplicity; those based on image generation rely heavily on large training dataset; those based on optimization are time-consuming and require a high computational cost. 

To this end, we propose a novel image2image translation model that converts the input face image into its corresponding UV map. The proposed method is image generation-based,  therefore much faster than optimization-based methods. We use the pseudo UV map for training, bypassing the dependency on the real UV map dataset. Thanks to multiple partial UV discriminators, we can use cropped parts of incomplete UV maps (acquired using the data pre-processing method provided in~\cite{deng2018uv}) for training to improve the generation quality.  Our contributions are as follows: 

\begin{itemize}
    \item A novel image generation-based UV map prediction framework is proposed. The generated results are comparable to the optimization-based method but much faster. 
    \item With the proposed UV sampler module, the visible face textures can be directly mapped to the UV space, forming an incomplete UV map. No 3D information (shape, occlusion) is required during the inference stage. Therefore our model can be stitched seamlessly with any 3D shape reconstruction models.
    \item The training is doesn't rely on the real UV map dataset, and the design of multiple discriminators can compensate well for the imperfect ground truth. 
    \item The proposed method outperforms the state-of-the-art methods, both qualitatively and quantitatively. 
\end{itemize}


\section{Related Work}

\textbf{3D shape reconstruction} From earlier optimization-based methods to CNN prediction-based methods, acquiring accurate 3D face shape becomes easier and faster, bringing powerful tools and significant opportunities for face-related tasks. Our training process relies on 3D shape reconstruction of a given face, where numerous 3D shape fitting methods are applicable. Since we train the model on large amounts of face images, the fitting speed is also crucial. In this paper, we adopt 3DDFA-V2 ~\cite{guo2020towards} as our shape re-constructor. Based on the backbone of MobileNet~\cite{howard2017mobilenets}, the model is lightweight and super-fast. With a single face image as input, the model will predict its corresponding pose and 3DMM shape parameters. And the 3D shape is computed as:
\begin{equation}
    \label{eq1}
    \mathbf{S} = \overline{\mathbf{S}}+\mathbf{A}_{id}\bm{\alpha}_{id}+\mathbf{A}_{exp}\bm{\alpha}_{exp}
\end{equation}
where $\mathbf{S}$ is the fitted 3D face shape, $\overline{\mathbf{S}}$ is the mean 3D shape, $\mathrm{A}_{id}$ and $\mathrm{A}_{exp}$ are the identity and the expression bases of 3DMM~\cite{paysan20093d}. $\bm{\alpha}_{id}\in\mathbb{R}^{40}$ and $\bm{\alpha}_{exp}\in\mathbb{R}^{10}$ are the predicted identity/expression parameters corresponding to $\mathbf{A}_{id}$ and $\mathbf{A}_{exp}$, respectively. After 3D shape reconstruction, $\mathbf{S}$ can be transformed and projected onto the image plane as follows:
\begin{equation}
    \label{eq2}
    V_{2d}(\mathbf{p}) = \mathbf{Pr}\cdot(\mathbf{R}\cdot\mathbf{S}+\mathbf{t}_{\mathrm{3d}})
\end{equation}
where $\mathbf{Pr}$ is the orthogonal projection matrix from 3D to 2D, $\mathbf{R}\in\mathbb{R}^{3\times 3}$ and $\mathbf{t}_{\mathrm{3d}}\in\mathbb{R}^{3\times 1}$ are the predicted pose parameters. 

\textbf{UV map generation} There exist mainly two texture representation methods for 3D models, vertex-based and UV map-based. The vertex-based representation is very intuitive, where each vertex has a color, and the interpolation of those colors generates the texture of the 3D surface. However, such representation flattens the texture into a linear vector, destroys the spatial relationship of texture patches, thus prevents it from leveraging powerful CNN-based methods. The UV map-based representation unwraps the 3D texture into a 2D space. Briefly, each 3D vertex's color is mapped to its corresponding location of a 2D image, and adjacent vertices are mapped to adjacent regions so that the positional relationships between vertices are well preserved.~\cite{deng2018uv} use Equation ~\ref{eq2} to sample the color of visible 3D vertices, then map them to UV space to get the incomplete UV map, in which the missing parts will be further completed by the generative model. However, their method is highly dependent on the precise 3D shape and ground truth UV maps. In contrast, our method does not need the UV map data for training or 3D shape for inference.~\cite{tran2018nonlinear} propose a non-linear 3DMM, where the predicted texture takes the UV map-based representation. Nevertheless, their UV map generator's input is a low-dimensional encoding of the input image, resulting in an 
loss of detail of the predicted UV map. In addition, their model is trained on linear 3DMM synthesized images~\cite{zhu2016face}, where artifacts caused by self-occlusion appear frequently. Unlike ~\cite{tran2018nonlinear}, our model is trained on real face images, and the coding keeps a large dimension across the forward path, making the generated UV photorealistic.  

\textbf{Differentiable renderer} To obtain the gradient of the loss function and thus train the network, a differentiable renderer is widely used in 3D face-related algorithms~\cite{richardson2017learning,guo2018cnn,tran2018nonlinear,deng2019accurate}. Briefly, a renderer is composed of a rasterizer and a shader. The rasterizer applies depth-buffering to select the mesh triangles corresponding to each pixel, and the shader computes the pixel colors as follows:
\begin{equation}
    \label{eq3}
    \bar{c} = w_0 c_0 + w_1 c_1 + w_2 c_2
\end{equation}
where $c_i$ is the color of the $i^{th}$ vertex of the mesh triangle the pixel resides in, $w_i$ is the barycentric coordinate of the pixel in the triangle. During backward propagation, the gradients are passed from each pixel to the vertices:
\begin{equation}
    \label{eq4}
    \frac{\mathrm{d}L}{\mathrm{d}c_i}=\frac{\mathrm{d}L}{\mathrm{d}\bar{c}}\frac{\mathrm{d}\bar{c}}{\mathrm{d}c_i}=\frac{\mathrm{d}L}{\mathrm{d}\bar{c}}w_i
\end{equation}
where $L$ is the loss function. Since $c_i$ is sampled from the output of the texture generator, \textit{i.e.}, the UV map, the gradients could be further backpropagated. In our project, we adopt the off-the-shelf differentiable renderer of PyTorch3D ~\cite{ravi2020pytorch3d}.

\textbf{Pixel attention sampling} To get the UV map of visible parts, UV-GAN ~\cite{deng2018uv} first fits a 3DMM to the input image, then uses Equation~\ref{eq2} to get the corresponding pixel location of each visible vertices, vertices' colors could be sampled, and the incomplete UV map is further generated. However, their method relies on accurate 3D shape fitting and facial landmark detection. Inspired by ~\cite{yin2020pixel}, we apply a pixel attention sampling (PAS) module to sample the incomplete UV map from the input image directly. Thanks to this module, the inference process is free from 3D shape or facial landmarks. Besides, different from ~\cite{yin2020pixel}, where input images require landmark-based pre-alignment due to the arbitrary target poses. The target output, \textit{i.e.}, the UV map, is highly structured, so neither spatial transformation to the input image nor the target pose condition is necessary.

\section{Proposed Method}

The goal of our method is to predict the face UV map from a single face image. As illustrated in Figure~\ref{fig:fig2}, the proposed model consists of two parts: a UV attention sampling module (UV sampler) and a UV map inpainting module (UV generator). During the inference process, the UV sampler will sample the pixels from the input image to generate an incomplete UV map, and then the UV generator will further complete the semi-finished UV map.  We describe the details of each component as follows. 

\subsection{UV Attention Sampling}

\begin{figure}[t]
\begin{center}
  \includegraphics[width=0.8\linewidth]{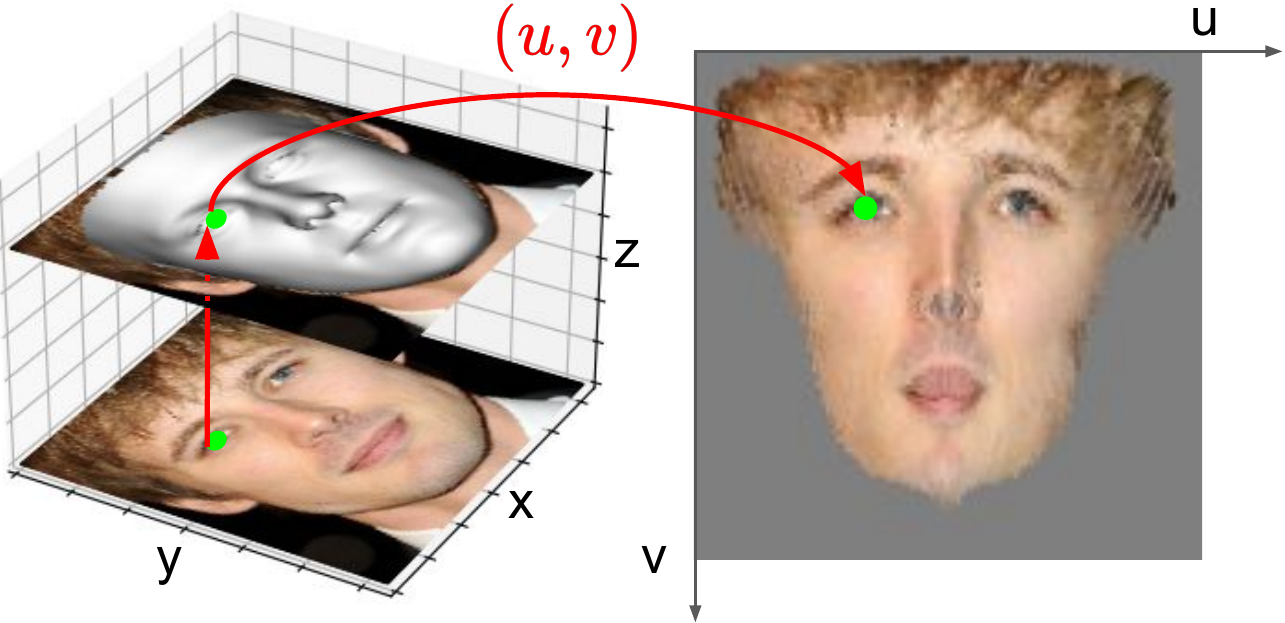}
\end{center}
  \caption{The traditional method for incomplete UV map generation. Which is used for generating the target output of the UV sampler.}
\label{fig:fig3}
\end{figure}

\begin{figure*}[t]
\begin{center}
\includegraphics[width=0.8\textwidth]{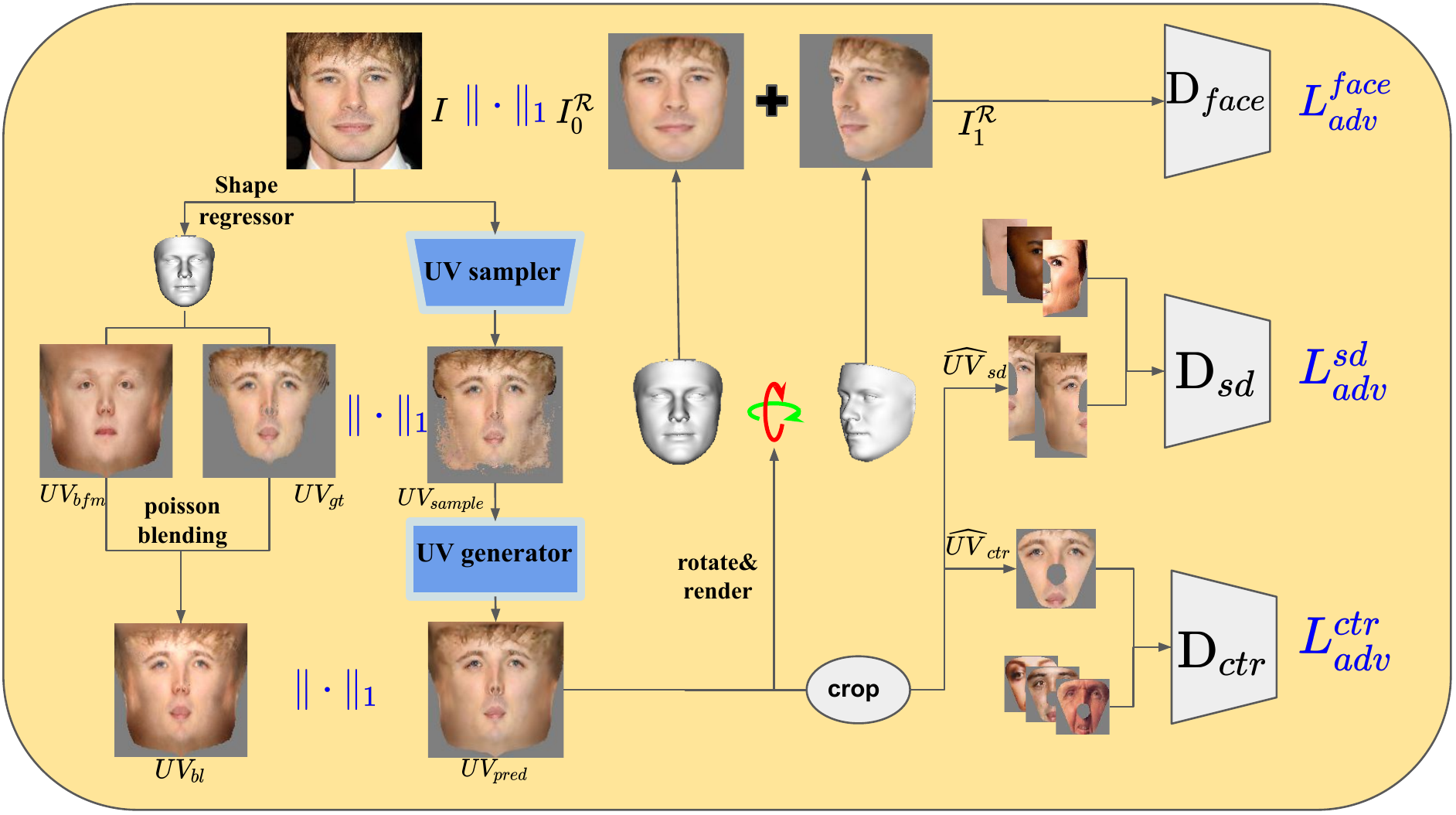}
\end{center}
  \caption{Overview of our approach. (1) Given an input image $I$, the UV sampler samples the visible textures and generate an incomplete UV map $UV_{sample}$ (2) The UV generator will further complete the sampled UV map and output the $UV_{pred}$. (3) With an off-the-shelf shape regressor, we predict the shape of the input face image, which is used for getting the ground truth of the $UV_{sample}$: $UV_{gt}$ and the ground truth of the $UV_{pred}$: $UV_{bl}$. (4) The predicted UV map is used to render face images of different poses: $I_0^{\mathcal{R}}$ and $I_1^{\mathcal{R}}$, which are further fed into a face discriminator. (5) $UV_{pred}$ is cropped to the side part $\widehat{UV}_{sd}$ and center part $\widehat{UV}_{ctr}$, fed into their corresponding discriminators.}
\label{fig:fig2}
\end{figure*}

The UV map is a two-dimensional representation of the global texture of a 3D object. Due to self-occlusion, it is an ill-posed problem to get the UV map from a single image. This section studies how to generate an incomplete UV map that contains only visible textures of the input face image. As a comparison, we recall the traditional method, which consists of four steps: (1) Get the 3D face shape based on the input image. (2) Determine the visible vertices using depth-buffer-based methods. (3) Project these visible vertices onto the image plane and index their colors according to their coordinates. (4) Render the UV map with the colors and the pre-defined UV-coordinates corresponding to each visible vertex. Figure~\ref{fig:fig3} illustrates the above steps. Obviously, such a method is tedious and relies on an accurate 3D shape fitting. Since the UV map contains \textit{only} the texture information of a 3D surface, is it really necessary to fit the exact 3D shape before getting the UV map? We do not think so. In fact, the only purpose of the 3D shape is to establish a one-to-one relationship between the pixel in the 2D face image and the pixel in the UV map, so why not learn such a mapping relationship in a data-driven manner? To achieve such a goal, we designed the UV sampler, a CNN-based model that maps the face image's pixels directly to the UV map. 

The model has two parts, \textit{i.e.}, the feature extractor and the sampler head. Similar to most generative models, the feature extractor is composed of stacked residual blocks~\cite{he2016deep}. Spectral normalization~\cite{miyato2018spectral} is applied to each convolution layer to stabilize the training. With this module, a 512-dimensional feature is extracted from the input image. Then, the feature is fed into the sampler head, a stack of fully connected layers interspersed with ReLU activations.  The sampler's output is reshaped as $S_{att} \in \mathbb{R}^{B\times 152\times 152\times 2}$, which is the attention sampling map, where $B$ is the batch size, 152 is the height/width of the UV map, and the last two channels hold the normalized abscissa and ordinate of the pixel in the input image to sample. Based on $S_{att}$, differentiable sampling~\cite{jaderberg2015spatial} is applied to the input image $I$, and an incomplete UV map $\widehat{UV}_{spl}$ is finally obtained.  

\begin{figure}[t]
\centering
\begin{subfigure}[b]{0.11\textwidth}
    \centering
    \includegraphics[width=\textwidth]{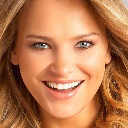}
    \caption{}
\label{fig:fig4a}
\end{subfigure}\hfill
\begin{subfigure}[b]{0.11\textwidth}
    \centering
    \includegraphics[width=\textwidth]{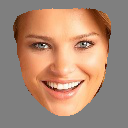}
    \caption{}
\label{fig:fig4b}
\end{subfigure}\hfill
\begin{subfigure}[b]{0.11\textwidth}
    \centering
    \includegraphics[width=\textwidth]{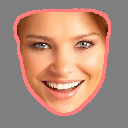}
    \caption{}
\label{fig:fig4c}
\end{subfigure}\hfill
\\
\begin{subfigure}[b]{0.11\textwidth}
    \centering
    \includegraphics[width=\textwidth]{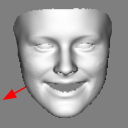}
    \caption{}
\label{fig:fig4d}
\end{subfigure}\hfill
\begin{subfigure}[b]{0.11\textwidth}
    \centering
    \includegraphics[width=\textwidth]{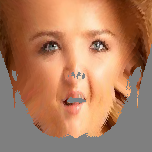}
    \caption{}
\label{fig:fig4e}
\end{subfigure}\hfill
\begin{subfigure}[b]{0.11\textwidth}
    \centering
    \includegraphics[width=\textwidth]{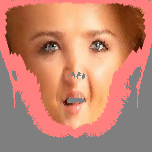}
    \caption{}
   \label{fig:fig4f}
\end{subfigure}\hfill

\caption{The impact of imprecise edge on UV map. \textbf{(a)} The input image. \textbf{(b)} The face cropped by the imprecise mask. \textbf{(c)} The face cropped by the eroded mask, where the red contour remarks the discarded edge pixels. \textbf{(d)} The 3D shape of the input face, where the red arrow represents the surface normal at the edge, which is nearly parallel to the image plane. \textbf{(e)} Inaccurate UV map generated from (b). \textbf{(f)} UV map generated from (c), the red part shows the area affected by edge pixels.}
\label{fig:fig4}
\end{figure}

To train the model, we use the above-mentioned traditional method to generate the ground truth (incomplete) UV map, $UV_{gt}$. One issue to note is that the 3D shape fitting is not completely accurate. The projected 3D vertices sometimes could lie on the background part of the face image, resulting in wrong vertices' colors and further wrong $UV_{gt}$. Moreover, since the surface normal of the face edge is almost parallel to the image plane, numerous projected vertices are piled up in the narrow edge region, aggravating the UV map's inaccuracy. To mitigate this problem, we first generate an \textit{imprecise} binary mask for the face region based on the projected 3D vertices, then erode the mask's edge to ensure that the region inside of which must be the face, the generation of $UV_{gt}$ only takes into account the vertices that fall inside the mask. Although this would result in a loss of texture near the edge, it is worth sacrificing the unimportant edges to ensure the accuracy of $UV_{gt}$. Figure~\ref{fig:fig4} illustrates the above discussion.

The training is guided by the following loss function:
\begin{equation}
    \label{eq5}
    \mathcal{L}_{spl} = \|\widehat{UV}_{spl}-UV_{gt}\|_1 +\|\hat{I}^{\mathcal{R}}-I^m\|_1+ \lambda TV(\widehat{UV}_{spl})
\end{equation}
where $I$ is the background masked input face image, $\widehat{UV}_{spl}$ is the output of the UV sampler, $\hat{I}^{\mathcal{R}}$ is the image rendered from the predicted texture and the 3D shape/pose ($S$/$P_0$) of the input face.
\begin{equation}
    \label{eq6}
    \hat{I}^{\mathcal{R}} = \mathcal{R}(\widehat{UV}_{spl}, S, P_0)
\end{equation}
$TV(\widehat{UV}_{spl})$ is the total variation loss~\cite{mahendran2015understanding} of the predicted UV map, which is powerful in smoothing the noises of the generated UV map.
\begin{equation}
\begin{aligned}
\label{eq7}
    &TV(\widehat{UV}_{spl}) =\\
     &\sum_{x,y,c=1}^{W-1,H,C}\left|\widehat{UV}_{spl}(x+1,y,c)-\widehat{UV}_{spl}(x,y,c)\right|^2+ \\
      &\sum_{x,y,c=1}^{W,H-1,C}\left|\widehat{UV}_{spl}(x,y+1,c)-\widehat{UV}_{spl}(x,y,c)\right|^2
\end{aligned}
\end{equation}

Thanks to the UV sampler, an incomplete UV map could be sampled directly from the input image, bypassing a series of complex and expensive steps of traditional methods, including 3D shape fitting, visible vertices determination, UV map rendering, etc.

\subsection{UV Map Inpainting}
With the UV sampler described above, we can sample an incomplete UV map from a face image. The next task is to fill the missing parts with textures consistent with sampled parts. This is an image inpainting problem, which has been extensively studied in the existing literature. However, most of the image inpainting methods are trained on paired images, which means that the ground truth image is uniquely determined, while in our UV map inpainting, the ground truth is not available. This section studies how to train a UV map inpainting model without the supervision of the ground truth. Briefly, our approach is to generate a pseudo ground truth UV map to assist the training. Then, we work with multiple discriminators to make the generated images as photorealistic as possible. 
\subsubsection{Pseudo UV Map Generation}
Generating the pseudo UV map consists of three steps: 1) incomplete ground truth UV map generation, 2) 3DMM texture fitting, 3) seamless image blending.  The first step has been described in detail in the previous section. Here we focus on the second and third steps. 

\begin{figure}[t]
\captionsetup[subfigure]{labelformat=empty}
\centering
\begin{subfigure}[b]{0.11\textwidth}
    \centering
    \includegraphics[width=\textwidth]{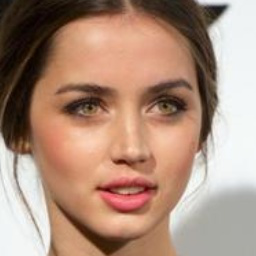}
    \caption{(a)}
\label{fig:fig5a}
\end{subfigure}\hfill
\begin{subfigure}[b]{0.11\textwidth}
    \centering
    \includegraphics[width=\textwidth]{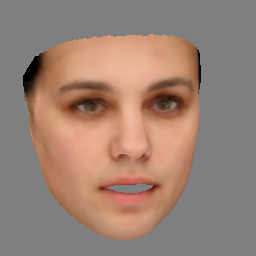}
    \caption{(c)}
\label{fig:fig5c}
\end{subfigure}\hfill
\begin{subfigure}[b]{0.11\textwidth}
    \centering
    \includegraphics[width=\textwidth]{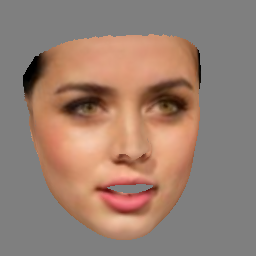}
    \caption{(e)}
\label{fig:fig5e}
\end{subfigure}\hfill
\begin{subfigure}[b]{0.11\textwidth}
    \centering
    \includegraphics[width=\textwidth]{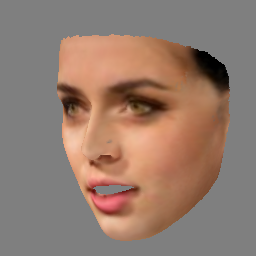}
    \caption{(g)}
\label{fig:fig5g}
\end{subfigure} \\
\begin{subfigure}[b]{0.11\textwidth}
    \centering
    \includegraphics[width=\textwidth]{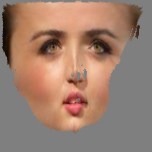}
    \caption{(b)}
\label{fig:fig5b}
\end{subfigure}\hfill
\begin{subfigure}[b]{0.11\textwidth}
    \centering
    \includegraphics[width=\textwidth]{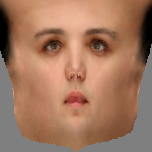}
    \caption{(d)}
\label{fig:fig5d}
\end{subfigure}\hfill
\begin{subfigure}[b]{0.11\textwidth}
    \centering
    \includegraphics[width=\textwidth]{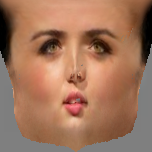}
    \caption{(f)}
   \label{fig:fig5f}
\end{subfigure}\hfill
\begin{subfigure}[b]{0.11\textwidth}
    \centering
    \includegraphics[width=\textwidth]{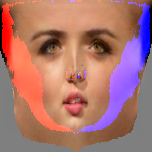}
    \caption{(h)}
\label{fig:fig5h}
\end{subfigure}
\caption{\textbf{(a)} The input image. \textbf{(b)} $UV_{gt}$. \textbf{(c)} The face reconstructed from $UV_{bfm}$. \textbf{(d)} $UV_{bfm}$. \textbf{(e)} The face reconstructed from $UV_{bl}$. \textbf{(f)} $UV_{bl}$. \textbf{(g)} The face in (e) under different view angle. \textbf{(h)} The texture marked in red is used to fill the missing texture in its symmetric area (marked in blue).}
\label{fig:fig5}
\end{figure}

\textbf{3DMM texture fitting} 
After the first step, the colors of visible 3D vertices (edge vertices excluded) are determined, with which we can get the 3DMM texture parameter by solving an optimization problem:
\begin{equation}
    \label{eq8}
    \min\limits_{p}\mathcal{E}= \|c\odot m - \mathrm{W}p\odot m\|^2_2+\lambda \|p\|^2_2
\end{equation}
where $c$ stores the colors of visible vertices, $m$ is the binary mask of visible vertices, $p$ is the texture parameter,  $\mathrm{W}$ is the texture basis of the 3DMM, more precisely BFM~\cite{bfm09}. Since the number of visible vertices is much more than the number of vectors in $\mathrm{W}$, Equation~\ref{eq8} defines an over-determined system, which has a unique solution. We take the UV map representation of the solved texture, denoted as $UV_{bfm}$. However, due to the linear, low-dimensional nature of the 3DMM, textures of $UV_{bfm}$ is far from reality, as can be seen in Figure~\ref{fig:fig5c}. Therefore, we move to the next step: seamless image blending. 

\textbf{Seamless image blending} The $UV_{gt}$ obtained in the first step can faithfully restore the input image (face edge excluded), but it is incomplete due to self-occlusion. While the $UV_{bfm}$ obtained in the second step is complete, but it is only a rough approximation in the 3DMM solution space. Thanks to Poisson image editing~\cite{perez2003poisson}, we can seamlessly blend the two results by solving the following Poisson equation with Dirichlet boundary conditions:
\begin{equation}
\label{eq9}
\Delta f = \mathrm{div}\ \mathbf{v} \  \text{over} \ \Omega,\  \text{with}\ f|_{\partial\Omega} = f^*|_{\partial\Omega}
\end{equation}
where in our case, $\Omega$ denotes the domain of real textures in $UV_{gt}$, $f$ is the texture to be modified in the $UV_{bfm}$, $\mathbf{v}$ is the gradient of the texture in $UV_{gt}$, $f^*$ denote the texture of $UV_{bfm}$ outside the $\Omega$. Besides, with the symmetry of the face UV map, we leverage the texture of the visible region to fill its occluded symmetric region, as illustrated in Figure~\ref{fig:fig5h}. This step is also based on Equation~\ref{eq9}. The final blending result is denoted as $UV_{bl}$, as in Figure~\ref{fig:fig5d}, both Figure~\ref{fig:fig5b} and Figure~\ref{fig:fig5f} are generated from it, which is far more photorealistic than the face generated from $UV_{bfm}$ in Figure~\ref{fig:fig5e}. 

\subsubsection{Multiple discriminator}
As with most generative models, the training of UV generator follows the adversarial paradigm, \textit{i.e.}, one or multiple discriminators are trained together with the generator. To train the discriminator, a large amount of data in the target domain is essential. However, the pseudo UV map $U_{bl}$ generated above is not very reliable. Its quality depends on the accuracy of $UV_{bfm}$, the area of valid texture in $UV_{gt}$, and the accuracy of the 3D shape. We only use the pseudo UV map to calculate the reconstruction loss, which is a rough guide to the generator's output. Although the complete UV map data is not available, we might as well collect a bunch of partial UV maps using the traditional method, \textit{i.e.}, for UV maps generated from frontal face images, the central region, denoted as $UV_{ctr}$, is accurate, and for UV maps generated from profile face images, the visible half side, $UV_{sd}$ is precise. Since the fitted 3D shape of the nose is not accurate enough, the nostril of $UV_{ctr}$ and the nose of $UV_{sd}$ are masked. Note that the partial UV maps collected in this way are not paired with $UV_{pred}$, so they are only used to train discriminators, thus indirectly force the $UV_{pred}$ lying in the real domain.  

\begin{figure}[t]
    \centering
    \includegraphics[width=0.8\linewidth]{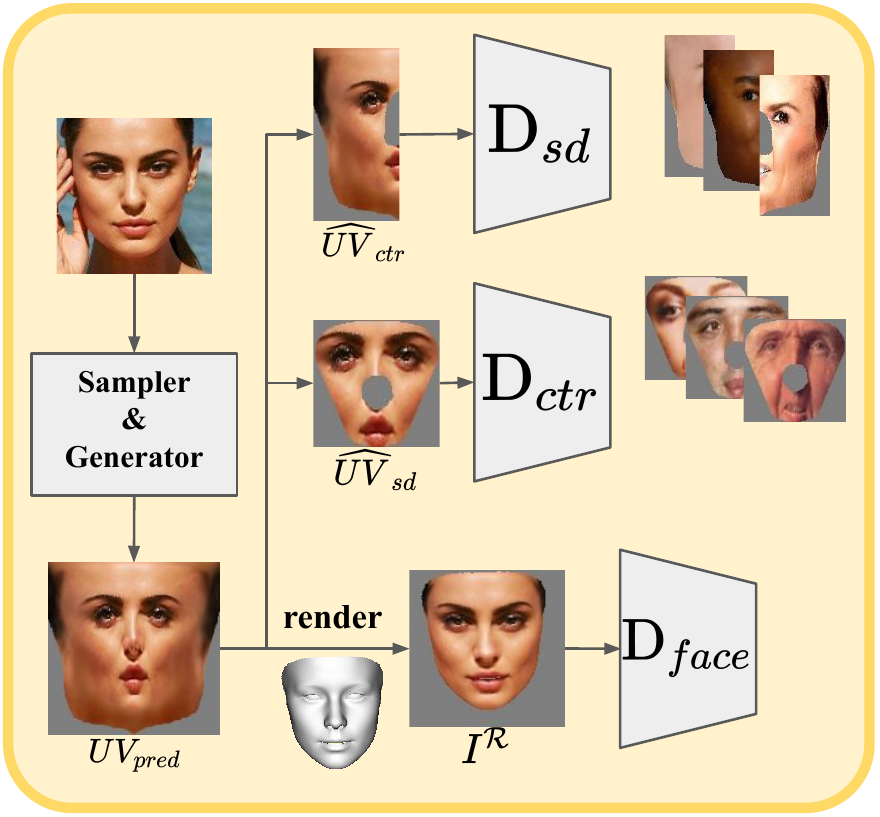}
    \caption{Three discriminators of our system, $D_{sd}$ is the side UV discriminator, $D_{ctr}$ is the center UV discriminator, $D_{face}$ is the face discriminator. At the end of $D_{ctr}$ and $D_{sd}$ shows the real data for that discriminator.}
    \label{fig:fig6}
\end{figure}

We design two partial UV map discriminators, one for the half side, the other for the center region. Together with the \textit{cropped face} discriminator, the system has three discriminators in total, as shown in Figure~\ref{fig:fig6}. 

We train our UV generator using the following losses.

\textbf{Adversarial loss} Given an output of the UV sampler, $UV_{spl}$, the generator will predict a global UV map, $UV_{pred}$. Three UV patches can be cropped from $UV_{pred}$, namely $\widehat{UV}_{ctr}$, $\widehat{UV}_{left}$, $\widehat{UV}_{right}$. Due to UV map's symmetry, the latter two can be put together and denoted as $\widehat{UV}_{sd}$. With the $UV_{pred}$ and the 3D shape/pose parameters predicted by the 3DDFA~\cite{guo2020towards} model, a reconstructed face image $I_0^{\mathcal{R}}$ could be rendered. By changing the pose parameter, we can get a face image in a different view angle, denoted as $I_1^{\mathcal{R}}$. So far, we have three types of fake data: $\widehat{UV}_{sd}$, $\widehat{UV}_{ctr}$ and $I_{0,1}^{\mathcal{R}}$, each of which corresponds to real data represented as $UV_{sd}$, $UV_{ctr}$ and $I^m$, where $I^m$ is the input face image with background masked.  

The adversarial loss is thus formulated as:
\begin{equation}
    \label{adv}
        \mathcal{L}_{adv} = \mathbb{E}_{x}[\mathrm{log}D(x)]+\mathbb{E}_{\hat{x}}\mathrm{log}(1-D(\hat{x}))
\end{equation}
where
\begin{equation*}
\begin{aligned}
        (x, \hat{x}, D)\in \{&(UV_{ctr}, \widehat{UV}_{ctr}, D_{ctr}), \\
        &(UV_{sd}, \widehat{UV}_{sd}, D_{sd})\\
        &(I^m, I_{0,1}^{\mathcal{R}}, D_{face})
        \}
\end{aligned}
\end{equation*}

\textbf{Reconstruction loss} The reconstruction loss consists of two terms, the UV reconstruction loss and the face reconstruction loss. 
\begin{equation}
    \label{UVrecon}
    L_{rec} = \|UV_{pred}-UV_{bl}\|_1 + \|I^{\mathcal{R}}-I^m\|_1
\end{equation}

\textbf{Symmetry loss} Since the UV map of the face is left-right symmetrical, we design the symmetry loss to help the model learn this property.
\begin{equation}
    L_{sym} = \|UV_{pred}-FlipLR(UV_{pred})\|_1
\end{equation}

\textbf{Identity loss} 
Since the pose is arbitrary, the ground truth of $I_1^{\mathcal{R}}$ is not available. Thus we use the pre-trained LightCNN~\cite{wu2018light} to extract the identity feature of $I^{\mathcal{R'}}$ and $I^M$, and minimize their $L_1$ distance. 
\begin{equation}
    \label{loss_id}
    L_{id} = \|\mathcal{F}(I^{\mathcal{R}'})-\mathcal{F}(I^m)\|_1
\end{equation}

\textbf{TV loss} TV loss of Equation~\ref{eq7} is also applied to $UV_{pred}$. 

The total loss function is as follows:
\begin{equation}
    L = L_{rec} + \lambda_1 L_{adv} + \lambda_2 L_{sym} + \lambda_3L_{id}+\lambda_4 TV
\end{equation}

\section{Experiments}
The proposed method can faithfully convert the input face image to its corresponding UV map. To demonstrate the conversion ability, we qualitatively compare the 3D reconstruction results with the current state-of-the-art methods, both 2D-based and 3D-based. A quantitative evaluation is also presented. 

\subsection{Implementation details}
Our training is based on three datasets: CelebA~\cite{liu2015deep}, CelebA-HQ~\cite{karras2017progressive}, and FFHQ~\cite{karras2019style}. Face images are pre-aligned with landmarks detected by~\cite{bulat2017far}. The input image size is $256\times 256$, and the predicted UV map is the same size as the input. We set the learning rate to $1e-4$ and use Adam\cite{kingma2014adam} optimizer with betas of [0.5, 0.999], the batch size is set to 6. We first pre-train the UV sampler until it outputs an incomplete UV map that can perfectly reconstruct the input image, which takes about 30000 steps. Then we train the UV generator for 100K steps with the UV sampler's weights fixed. We found that directly training the sampler to produce $256\times 256$ UV map does not converge easily, resulting in a large amount of noise. Since the UV sampler actually predicts the \textit{normalized} abscissa and ordinates of the pixels in the input image to sample, we could first train it on 128 sized images to generate 152 sized UV maps, then apply a simple bilinear interpolation of the predicted sampling coordinates and use them to sample the 256 sized images (fine-tuning needed). The training of the whole model takes about 100 hours on two Titan X Pascal graphics cards. Since most of the face images in the training sets are frontal, making the model not robust to the large view angles, to solve this problem, one trick we adopt is to rotate and render the input faces with their corresponding shapes and pseudo UV maps, then train the model to reconstruct the original face images. 

\subsection{Qualitative results}
We use the predicted UV maps to render 3D shapes. By changing the pose parameters, images of different view angles are generated. For the qualitative evaluation, as a usual convention, we take the same inputs as others and paste the generated results after them.
\begin{figure}[t]
    \centering
    \includegraphics[width=\linewidth]{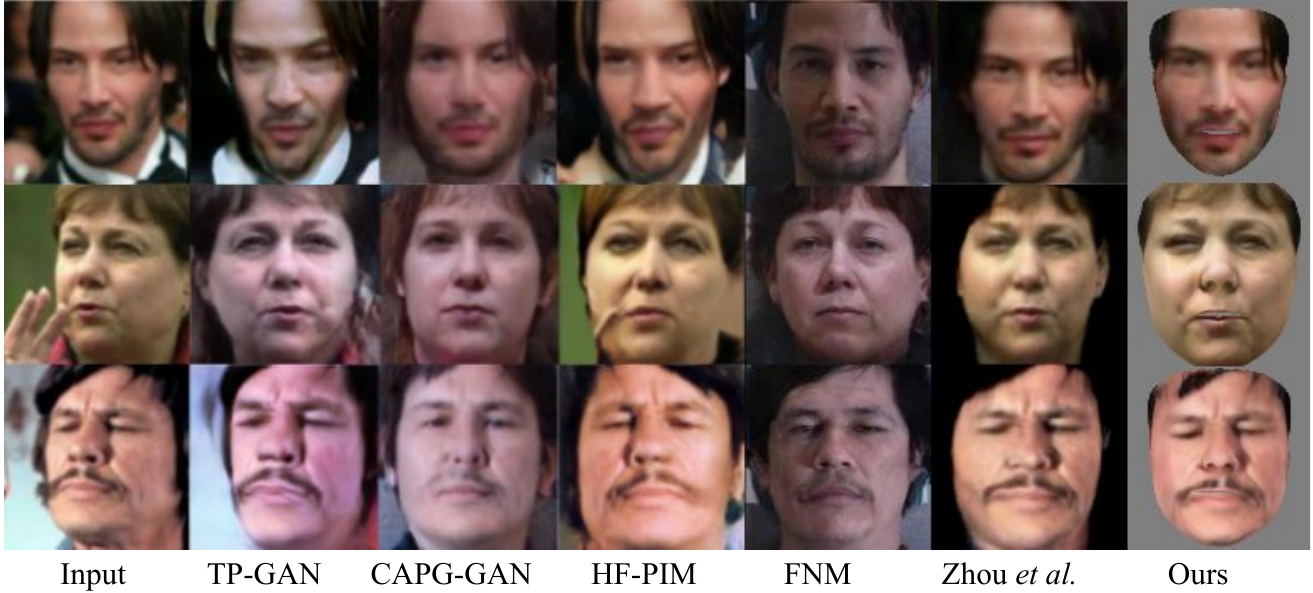}
    \caption{Frontalization results comparing with 2D-based face pose editing methods. Zoom-in for a better view.}
    \label{fig:fig7}
\end{figure}
Figure~\ref{fig:fig7} compares our frontalization results with 2D-based face pose editing methods, including TP-GAN~\cite{Huang_2017_ICCV}, CAPG-GAN~\cite{hu2018pose}, HF-PIM~\cite{cao2018learning}, FNM~\cite{qian2019unsupervised} and Zhou\textit{el al.}\cite{zhou2020rotate}. As shown in Figure~\ref{fig:fig7}, TP-GAN doesn't convert the pose well, and the third face image it generates is obviously left-skewed. Furthermore, the images generated by TP-GAN, CAPG-GAN, and FNM have large color deviations with the input images due to the influence of Multi-PIE~\cite{gross2010multi} data in the training set. Besides our method, only HF-PIM and Zhou \textit{et al.} maintain a consistent texture style with the input image. However, due to the lack of a priori knowledge of the 3D shape, HF-PIM cannot well preserve the face shape while editing the face pose. Our method achieves similar performance to the current state-of-the-art, Zhou \textit{et al.}, we use the same off-the-shelf shape regressor. However, their method is based on face image generation, which means that we need to re-infer the missing texture each time we change the view angle. Another limitation of face image generation-based method is that the training settings greatly limit their pose editing freedom. The method of Zhou \textit{et al.} cannot well generate face images of a large yaw angles; TP-GAN, FNM, and HF-PIM can only generate face images in frontal view.

\begin{figure*}[t]
    \centering
    \includegraphics[width=\textwidth]{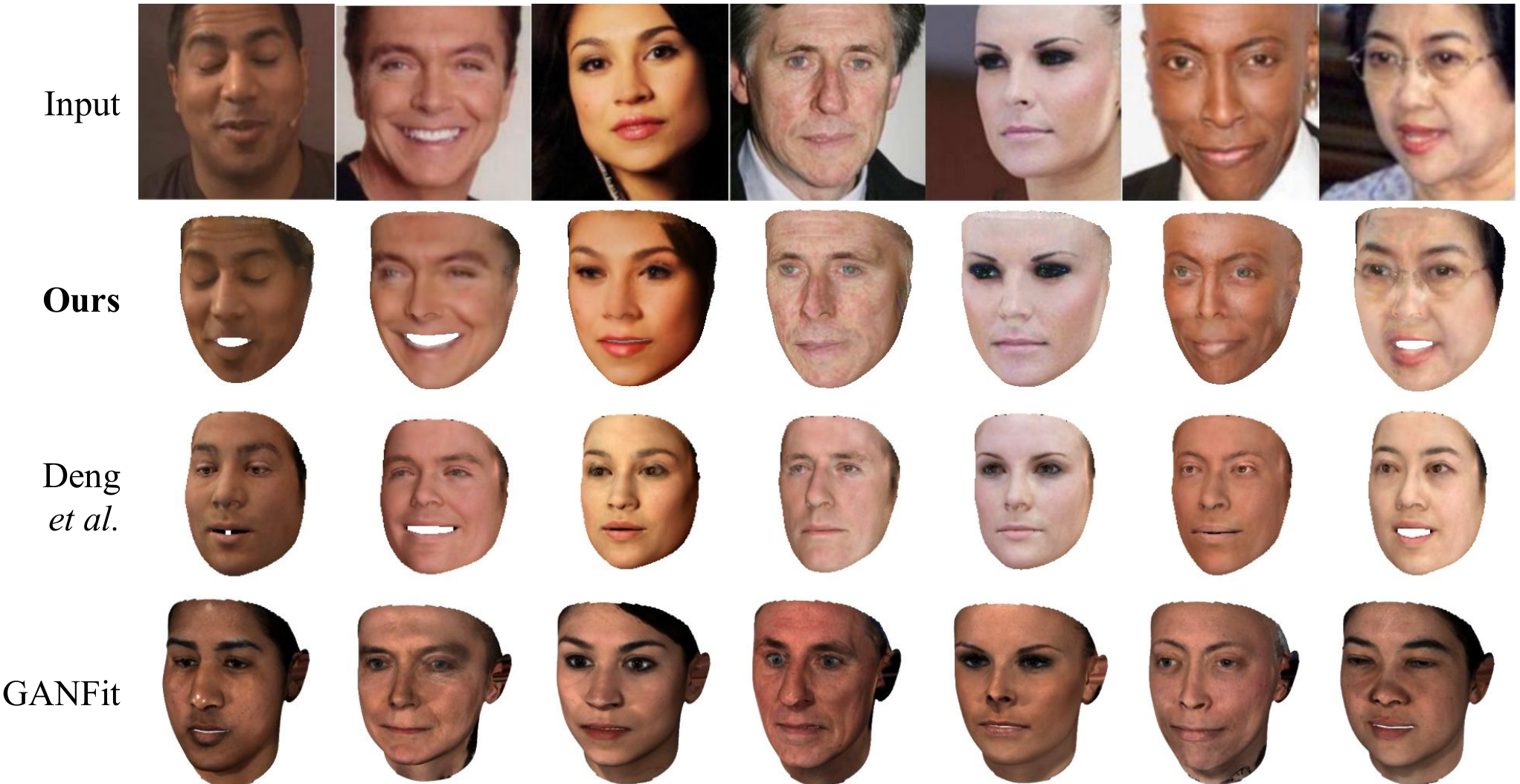}
    \caption{Qualitative comparison with other state-of-the art 3D reconstruction methods.}
    \label{fig:fig8}
\end{figure*}

A further qualitative comparison of our method and two representative 3D reconstruction-based methods are demonstrated in Figure~\ref{fig:fig8}. The method proposed by Deng \textit{et al.}~\cite{deng2019accurate} is based on 3DMM parameter regression, and GANFIT~\cite{gecer2019ganfit} is based on latent-code optimization of a pre-trained GAN model. As can be seen, our results are more visually pleasant: large amounts of details are well preserved, including freckles, wrinkles, and expressions. Due to the model's low-dimensional linear nature, it's difficult for 3DMM-based methods to restore the input image's details faithfully. As can be seen in the 3rd-row of~\ref{fig:fig8}, freckles and wrinkles are not well preserved. The results of GANFIT do contain richer details, but the resulting textures' styles are very homogeneous and differ considerably from their corresponding input images. We believe this is due to the lack of diversity in their training data, as the face UV map datasets are not easily accessible.

\subsection{Quantitative results}

\begin{table}[]
\caption{Comparison of the reconstruction and the identity-preserving ability on AFLW2000-3D, non-facial areas of all images are masked out for fair comparison.}
    \centering
    \begin{tabular}{|c| c| c| c |c|}
    \hline
    \ & \multicolumn{2}{c|}{Reconstruction} &\multicolumn{2}{c|}{Recognition}\\
    \hline
    Method & $L_{1}$& SSIM&Recon &Front\\
    \hline
    Deng \textit{et al.}& 0.064 & 0.698 & 0.554 &0.501\\
    \hline
    Zhou \textit{et al.}&0.069&0.613& 0.780& 0.675\\
    \hline
    \textbf{Ours}& \textbf{0.021}&\textbf{0.915} &\textbf{0.860} & \textbf{0.685}\\
    
    \hline

    \end{tabular}
    \label{tab:my_label}
\end{table}

We evaluate the proposed method in two aspects: the reconstruction ability and the identity-preserving ability. As most previous works are not open-sourced, we only compare with Deng \textit{et al.}~\cite{deng2019accurate} and Zhou \textit{et al.}~\cite{zhou2020rotate}, SOTA methods based on 3DMM and 2D face image generation, respectively. We conduct the experiments on the AFLW2000-3D~\cite{zhu2016face}, which contains 2000 face images with ground truth shape parameters. 

For the reconstruction ability evaluation, we calculate the L1 loss and the structural similarity~\cite{wang2003multiscale} of the reconstructed face images. As can be seen from Table~\ref{tab:my_label},  our method outperforms others in both these metrics. 

As for the identity-preserving ability, the evaluation is conducted by features extracted using the pre-trained LightCNN-29 v2~\cite{wu2018light} model. We calculate the cosine similarity of the features corresponding to the input images and the reconstructed/frontalized images. Results are shown in the two rightmost columns of Table~\ref{tab:my_label}. An interesting thing to notice is that, although Zhou \textit{et al.} is inferior to the 3DMM-based model in terms of reconstruction loss, they are more capable of preserving the face identity. However, our proposed method achieves the best performance in both aspects.



\section{Conclusion}
This work proposes a novel 2-stage image2image translation model that can convert the input face image into its corresponding UV map. In the first stage, with the proposed UV sampler, pixels in the input face images are selectively sampled and adjusted to form an incomplete UV map, which contains all the visible textures of the face. With the help of this module, the inference stage no longer requires the intervention of 3D shapes. In the second stage, the incomplete UV map is further completed by a UV generator. The training is conducted on purely pseudo UV maps, thus weakly-supervised. With the help of two carefully designed partial UV discriminators, we can generate photo-realistic face textures without the supervision of the complete UV map. Qualitative and quantitative experiments validate the reconstruction ability and the identity-preserving ability of the proposed method. 

{\small
\bibliographystyle{ieee_fullname}
\bibliography{egbib}
}

\end{document}